\documentclass[a4paper, 12 pt, twocolumn, final, conference]{ieeetran}
\IEEEoverridecommandlockouts

\usepackage{graphicx}
\usepackage{subcaption}
\usepackage{times}   
\usepackage{amsmath} 
\usepackage{amssymb} 
\usepackage{avldefs} 
\usepackage{algorithm}
\usepackage{amsfonts}
\usepackage{algorithm, algorithmic}
\usepackage{array}
\usepackage{multirow}
\usepackage{microtype}
\usepackage{graphics}
\usepackage{booktabs} 
\usepackage{url}
\usepackage{bm}  

\usepackage{todonotes}

\usepackage[square, numbers, compress]{natbib}

\usepackage{color}

\usepackage{booktabs}
\usepackage[normalem]{ulem}
\useunder{\uline}{\ul}{}

\title{\LARGE \bf
Slow Momentum with Fast Reversion: \\A Trading Strategy Using Deep Learning and Changepoint Detection
}

\author{Kieran Wood, Stephen Roberts, Stefan Zohren
\thanks{K. Wood, S. Roberts and S. Zohren are with the Department of Engineering Science and the Oxford-Man Institute of Quantitative Finance, University of Oxford, Oxford, United Kingdom (email: kieran.wood@eng.ox.ac.uk, sjrob@robots.ox.ac.uk, zohren@robots.ox.ac.uk).}
}

\makeatletter
\renewcommand*{\thetable}{\arabic{table}}
\renewcommand*{\thefigure}{\arabic{figure}}
\let\c@table\c@figure
\makeatother 

\usepackage{caption}
\begin{document}
\maketitle
\thispagestyle{plain}
\pagestyle{plain}
%

\begin{abstract}
Momentum strategies are an important part of alternative investments and are at the heart of commodity trading advisors (CTAs). These strategies have, however, been found to have difficulties adjusting to rapid changes in market conditions, such as during the 2020 market crash. In particular, immediately after momentum turning points, where a trend reverses from an uptrend (downtrend) to a downtrend (uptrend), time-series momentum (TSMOM) strategies are prone to making bad bets. To improve the response to regime change, we introduce a novel approach, where we insert an online changepoint detection (CPD) module into a \textit{Deep Momentum Network}  (DMN) \cite{DeepMomentum} pipeline, which uses an LSTM deep-learning architecture to simultaneously learn both trend estimation and position sizing. Furthermore, our model is able to optimise the way in which it balances 1) a slow momentum strategy which exploits persisting trends, but does not overreact to localised price moves, and 2) a fast mean-reversion strategy regime by quickly flipping its position, then swapping it back again to exploit localised price moves. Our CPD module outputs a changepoint location and severity score, allowing our model to learn to respond to varying degrees of disequilibrium, or smaller and more localised changepoints, in a data driven manner. Back-testing our model over the period 1995--2020, the addition of the CPD module leads to an improvement in Sharpe ratio of one-third. The module is especially beneficial in periods of significant nonstationarity, and in particular, over the most recent years tested (2015--2020) the performance boost is approximately two-thirds. This is interesting as traditional momentum strategies have been underperforming in this period.
\end{abstract}


\section{Introduction}
Time-series (TS) momentum \cite{TimeSeriesMomentum} strategies are derived from the philosophy that strong price trends have a tendency to persist. These trends have been observed to hold across a range of timescales, asset classes and time periods \cite{CenturyOfTrendFollowing, TwoCenturiesOfTrendFollowing, AHLMomentum}. Momentum strategies are often referred to as `follow the winner', because it is assumed that winners will continue to be winners in the subsequent period. Momentum strategies are an important part of alternative investments and are at the heart of commodity trading advisors (CTAs). Much effort goes into quantifying the magnitude of trends \cite{AHLMomentum, WhichTrendIsYourFriend, TrendFilteringLyxor} and sizing traded positions accordingly \cite{TSMomAndVolScaling, DemystifyingTimeSeriesMomentum, VolTargeting}. Rather than using handcrafted techniques to identify trends and select positions, \cite{DeepMomentum} introduces \textit{Deep Momentum Networks} (DMNs), where a Long Short-Term Memory (LSTM) \cite{lstm} deep learning architecture achieves this by directly optimising  on the  Sharpe  ratio  of  the  signal. Deep Learning has been widely utilised for time-series forecasting \cite{TimeSeriesDeepLearning}, achieving a high level of accuracy across various fields, including the field of finance for both daily data \cite{DeepMomentum, LSTMFinancialTimeSeries,EmpiricalAssetPricingViaML, XSectionalDeepMomentum, LearningToRank} and in a high frequency setting, using limit order book data \cite{RamaContUniversality, DeepLOB}. In recent years, implementation of such deep learning models has been made accessible via extensive open source frameworks such as \texttt{TensorFlow} \cite{tensorflow} and \texttt{PyTorch} \cite{pytorch}.

Momentum strategies aim to capitalise on persisting price trends, however occasionally these trends break down, which we label momentum turning points. At these turning points, momentum strategies are prone to performing poorly because they are unable to adapt quickly to this abrupt change in regime. This concept is explored in \cite{MomentumTurningPoints} where a slow momentum signal based on a long lookback window, such as 12 months, is blended with a fast momentum signal which is based on a short lookback window, such as 1 month. This approach is a balancing act between reducing noise and being quick enough to respond to turning points. Adopting the terminology from \cite{MomentumTurningPoints}, a Bull or Bear market is when the two momentum signals agree on a long or short position respectively. If slow momentum suggests a long (short) position, and fast momentum a short (long) position, we term this a Correction (Rebound) phase.

Correction and Rebound phases, where the momentum assumption breaks down, are examples of mean-reversion \cite{DoesStockMarketOverreact, MeanReversion, SeasonalityMeanReversion} regimes.  Mean-reversion  trading strategies, often referred to as `follow the loser' strategies, assume losers (winners) over some lookback window will be winners (losers) in the subsequent period. If we observe the positions taken by a DMN, alongside exploiting persisting trends, the model also exploits fluctuations in returns data at a shorter time horizon by regularly flipping its position then quickly changing back again. We argue that the high Sharpe ratio achieved by DMNs can be largely attributed to its fast mean-reversion property.

Changepoint detection (CPD) is a field which involves the identification of abrupt changes in sequential data, where the generative parameters for our model after the changepoint are independent of those which come before. The nonstationarity of real world time-series in fields such as finance, robotics and sensor data has led to a plethora of research in this field. To enable us to respond to CPD in real time we require an `online' algorithm, which processes each data point as it becomes available, as opposed to `offline' algorithms which consider the entire data set at once and detect changepoints retrospectively. First introduced by \cite{BOCPD}, Bayesian approaches to online CPD, which naturally accommodate to noisy, uncertain and incomplete time-series data, have proven to be very successful. Assuming a changepoint model of the parameters, the Bayesian approach integrates out the uncertainty for these parameters as opposed to using a point estimate. Gaussian Processes (GPs) \cite{GPRegression, GPs}, which are collections of random variables where any finite number of which have joint Gaussian distributions, are well suited to time-series modelling \cite{GPTimeSeries}. GPs are often referred to as a Bayesian non-parametric model and have the ability to handle changepoints \cite{GPChangepointsAndFaults, GPCPModels, GPAutoConstructionNL}. Rather than comparing a slow and fast momentum signals to detect regime change, we utilise GPs as a more principled method for detecting momentum turning points. For our experiments, we use the \texttt{Python} package \texttt{GPflow} \cite{GPflow2017} to build Gaussian process models, which leverage the \texttt{TensorFlow} framework.

In this paper, we introduce a novel approach, where we add an online CPD module to a DMN pipeline, to improve overall strategy returns. By incorporating the CPD module, we optimise our response to momentum turning points in a data-driven manner by passing outputs from the module as inputs to a DMN, which in turn learns trading rules and optimises position based on some finance value function such as Sharpe Ratio \cite{SharpeRatio}. This approach helps to correctly identify when we are in a Bull or Bear market and select the momentum strategy accordingly. With the addition of the CPD module, the new model learns how to exploit, but not overreact, to noise at a shorter timescale. Our strategy is able to exploit the fast reversion we observe in DMNs but effectively balance this with a slow momentum strategy and improve returns across an entire Bull or Bear regime. Effectively the new pipeline has more knowledge on how to respond to abrupt changes, or lack of changes in a data driven way. 

We argue that the concept of CPD is an artificial construct which can occur at varying degrees of severity and is dependent on decisions such as length of the lookback horizon for CPD. Rather than specifying regimes based on some criteria or threshold, we use our CPD module to quantify, or score, the level of disequilibrium, allowing the model to consider smaller or more localised `regime changes'. The length of the lookback window is the most sensitive design choice for the CPD module, for if the lookback horizon is too long, we miss smaller, but still potentially significant regime changes. If the horizon is too short, the data becomes too noisy and is of little value. We introduce the lookback window length (LBW) as a structural hyperparameter which we optimise using the outer optimisation loop of our model. This allows the module to be more tightly coupled with our LSTM module, helping us to maximise the efficiency of the CPD, and allowing us to tweak the LSTM hyperparamters in conjunction with the LBW.

It can be noted that the performance of DMNs, without CPD, deteriorates in more recent years. The deterioration in performance is especially notable in the period 2015--2020, which exhibits a greater degree of turbulence, or disequilibrium, than the preceding years. One possible explanation of deterioration in momentum strategies in recent years is the concept of `factor crowding', which is discussed in depth in \cite{FactorCrowding}, where it is argued that arbitrageurs inflict negative externalities on one another. By using the same models, and hence taking the same positions, a coordination problem is created, pushing the price away from fundamentals. It is argued that momentum strategies are susceptible to this scenario. Impressively, the addition of a CPD module helps to alleviate the deterioration in performance and our model significantly outperforms the standard DMN model during the 2015--2020 period. A similar phenomenon can be observed from around 2003, when electronic trading was becoming more common, where the deep learning based strategies start to significantly outperform classic TSMOM strategies.

\section{Changepoint Detection Using Gaussian Processes}
A classic univariate regression problem, of the form $y(x)=f(x)+\epsilon$, where $\epsilon$ is an additive noise process, has the goal of evaluating the function $f$ and the probability distribution $p(y_*\vert x_*)$ of some point $y_*$ given some $x_*$. Our daily time-series data, for asset $i$, consists of a sequence of observations for (closing) price $\{p_t^{(i)}\}_{t=1}^T$, up to time $T$. Since financial time-series are nonstationary in the mean, for each time $t$ we take the first difference of the time-series, otherwise known as  the arithmetic returns,
\begin{equation}
    r_{t-1,t}^{(i)} = \frac{p_t^{(i)} - p_{t-1}^{(i)}}{p_{t-1}^{(i)}},
\end{equation}
in an attempt to remove linear trend in the mean.  Throughout this paper, for brevity, we will refer to $r_{t-1,t}$ simply as $r_{t}$. For the purposes of CPD, it is not computationally feasible, nor is it necessary, to consider the entire time-series, hence we consider the series $\{r^{(i)}_t\}^T_{t=T-l}$, with lookback horizon $l$ from time $T$. For every CPD window, where $\mathcal{T}=\{T-l, T-l+1,\ldots,T\}$, we standardise our returns as,
\begin{equation}
    \label{eqn:standardise-returns}
    \hat{r_t}^{(i)} = \frac{r_t^{(i)}-\mathbb{E}_\mathcal{T}\left[r_t^{(i)}\right]}{\sqrt{\mathrm{Var}_\mathcal{T}\left[r_t^{(i)}\right]}}.
\end{equation}
This step is taken for two reasons, we can assume that the mean over our window is zero and, with unit variance, we have more consistency across all windows when we run our CPD module.

Our approach to changepoint detection, involves a curve fitting approach for input-output pairs $(t, \hat{r}^{(i)}_t)$ via the use of  Gaussian Process (GP) regression \cite{GPs}. GP regression is a probabilistic, non-parametric method, popular in the fields of machine learning and time-series analysis \cite{GPTimeSeries}. It is a kernel based technique where the Gaussian Process $\mathcal{GP}$ is specified by a covariance function $k_\xi(\cdot)$, which is in turn parameterised by a set of hyperparameters $\xi$. In its common guise, the GP has a stationary kernel; however, it should be noted that GPs can readily work well even when the time-series is nonstationary \cite{NonStationaryGPTS}. We define the GP as a distribution over functions where,
\begin{equation}
    \hat{r}_{t}^{(i)} = f(t) +\epsilon_t, f\sim \mathcal{GP}(0,k_\xi), \epsilon_t \sim \mathcal{N}(0,\sigma^2_n),
\end{equation}
given noise variance $\sigma_n$, which helps to deal with noisy outputs which are uncorrelated.

It has been demonstrated in \cite{GPFinancial, GPVolForecasting} that a Matérn 3/2 kernel is a good choice of covariance function for noisy financial data, which tends to be highly non-smooth and not infinitely differentiable. This problem setting favours the least smooth of the Matérn family of kernels which is the 3/2 kernel. We parametrise our Matérn 3/2 kernel as,  
\begin{equation}
    k(x,x') = \sigma_h^2
    \left( 1 + \frac{\sqrt{3}|x - x'|}{\lambda} \right)
    e^{ \left(-\frac{\sqrt{3}|x - x'|}{\lambda}  \right)},
\end{equation}
with kernel hyperparameters $\xi_M=(\lambda, \sigma_h, \sigma_n)$, where $\lambda$ is the input scale and $\sigma_h$ the output scale. We define our covariance matrix, for a set of locations $\mathbf{x}=[x_1,x_2,\ldots x_n]$ as,
\begin{equation}
\label{eq:GPK}
\mathbf{K}(\mathbf{x},\mathbf{x}) = \left (
\begin{array}{cccc}
 k(x_1,x_1)  & \cdots & k(x_1,x_n)\\
 \vdots & \ddots & \vdots \\
 k(x_n,x_1) &  \cdots & k(x_n,x_n)
\end{array}
\right ).
\end{equation}
Using $\mathbf{\hat{r}}=[\hat{r}_{t-l},...,\hat{r}_{t}]$, we integrate out the function variables to give $p(\mathbf{\hat{r}}\vert \xi)=\mathcal{N}(\mathbf{0}, \mathbf{V})$, with $\mathbf{V}=\mathbf{K}+\sigma_n^2\mathbf{I}$. Since $p(\xi \vert \mathbf{\hat{r}})$ is intractable, we instead apply Bayes' rule, 
\begin{equation}
    p(\xi \vert \mathbf{\hat{r}}) = 
    \frac{p(\mathbf{\hat{r}}\vert \xi)p(\xi)}
    {p(\mathbf{\hat{r}})}
\end{equation}
and perform type II maximum likelihood on $p(\mathbf{\hat{r}}\vert \xi)$. We minimise the negative log marginal likelihood, 
\begin{equation}
    \label{eqn:nlml}
    \mathrm{nlml}_\xi = \min_\xi \left ( \frac{1}{2}\mathbf{\hat{r}}^\mathsf{T} \mathbf{V}^{-1}\mathbf{\hat{r}} + \frac{1}{2}\log |\mathbf{V}| + \frac{l+1}{2}\log 2\pi \right ).
\end{equation}
We use the \texttt{GPflow} framework to compute the hyperparameters $\xi$, which in turn uses the L-BFGS-B \cite{LBFGSB} optimisation algorithm via the \texttt{scipy.optimize.minimize} package.

In \cite{GPChangepointsAndFaults, GPTimeSeries} it is assumed that our function of interest is well behaved, except there is a drastic change, or changepoint, at $c\in\{t-l+1,t-l+2,\ldots, t-1\}$, after which all observations before $c$ are completely uninformative about the observations after this point. It is important to note that the lookback window (LBW) $l$ for this approach needs to be prespecified and it is assumed that it contains a single changepoint. Each of the two regions are described by different covariance functions $k_{\xi_1}$, $k_{\xi_2}$, in our case Matérn 3/2 kernels, which are parameterised by hyperparameters $\xi_1$ and $\xi_2$ respectively. The Region-switching kernel is,
\begin{equation}
\label{eqn:kregionswitching}
 k_{\xi_R}(x, x') = \left\{ \begin{array}{cc} 
                k_{\xi_1}(x, x') & x, x' < c \\
                 k_{\xi_2}(x, x') & x, x' \geq c \\
                0 & \mathrm{otherwise,} \\
                \end{array} \right.
\end{equation}
with full set of hyperparameters $\xi_R = \{\xi_1, \xi_2, c, \sigma_n \}$. Here, a changepoint can take multiple forms, with these cases being either a drastic change in covariance, a sudden change in the input scale, or a sudden change in the output scale. In the context of financial time-series we can think of these cases as either a change in correlation length, a change in mean-reversion length or a change in volatility. 

It is computationally inefficient to fit $2(l-1)$ GPs, to minimise $\mathrm{nlml}_{\xi_R}$ as in \eqref{eqn:nlml}, due to the introduction the discrete hyperparameter $c$. We instead borrow an idea from \cite{GPAutoConstructionNL} and approximate the abrupt change of covariance in \eqref{eqn:kregionswitching} using a sigmmoid function $\sigma (x) =1/\left(1+e^{-s(x - c)}\right)$ which has the properties $\sigma(x,x') = \sigma(x) \sigma(x')$ and $\bar{\sigma}(x,x') (1-\sigma(x)) (1-\sigma(x'))$. Here, $c\in (t-l, t)$ is the changepoint location and $s>0$ is the steepness parameter. Our Changepoint kernel is,
\begin{equation}
    \label{eqn:kregionswitchingapprox}
    k_{\xi_C}(x, x') = 
    k_{\xi_1}(x, x')  \sigma(x,x') +
    k_{\xi_2}(x, x') \bar{\sigma}(x,x'),
\end{equation}
with full set of hyperparameters $\xi_C = \{\xi_1, \xi_2, c, s, \sigma_n \}$. We can compute $\mathrm{nlml}_{\xi_C}$ by optimising the parameters a single GP, which is significantly more efficient than computing $\mathrm{nlml}_{\xi_R}$, despite having additional hyperparameters. This new kernel has the added benefit of capturing more gradual transitions from one covariance function to another, due to the additional of the steepness parameter $s$. We implement \eqref{eqn:kregionswitchingapprox} in \texttt{GPflow} via the \texttt{gpflow.kernels.ChangePoints} class, adding the constraint $c\in (t-l, t)$, which is not enforced by default. 

\begin{figure}[h]
\centering
\includegraphics[width=0.9\linewidth]{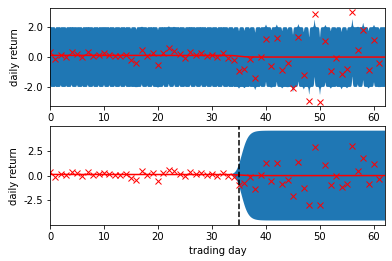}
\label{fig:cpd-hyp}
\caption{Plots of daily returns for S\&P 500, composite ratio-adjusted continuous futures contract during the first quarter of 2020, where returns have been standardised as per \eqref{eqn:standardise-returns}. The top plot fits a GP, using the Matérn 3/2 kernel and the bottom using the Changepoint kernel specified in \eqref{eqn:kregionswitchingapprox}. The shaded blue region covers $\pm2$ standard deviations from the mean and we can see that the top plot is dominated by the white noise term $\sigma_n\approx1$. The black dotted line indicates the location of the changepoint hyperparameter $c$ after minimising negative log marginal likelihood, which aligns with the COVID-19 market crash. The negative log marginal likelihood is reduced from $88.0$ to $47.9$, which corresponds to $\nu^{(i)}_t\approx 1$. }
\end{figure}

To quantify the level of disequilibrium, we look at the reduction in negative log marginal likelihood achieved via the introduction of the changepoint kernel hyperparameters, through comparison to $\textrm{nlml}_{\xi_M}$. If the introduction of additional hyperparameters leads to no reduction in negative log marginal likelihood, then the level of disequilibrium is low. Conversely, if the reduction is large, this indicates significant disequilibrium, or a stronger changepoint, because the data is better described by two covariance functions. Our changepoint score $\nu^{(i)}_t\in (0,1)$ and location $\gamma^{(i)}_t \in (0,1)$ are,
\begin{equation}
    \label{eqn:cpd-sev-and-loc}
    \nu^{(i)}_t = 1-\frac{1}{1+\mathrm{e}^{-(\mathrm{nlmn}_{\xi_C} - \mathrm{nlmn}_{\xi_M})}}, \quad \gamma^{(i)}_t = \frac{c - (t-l)}{l},
\end{equation}
which are both normalised values which helps to improve stability and performance of our LSTM module.

\section{Momentum Strategies Review}
\label{sec:mom-strats}
\subsection{Classical Strategies}
In this paper we focus on univariate time-series approaches \cite{TimeSeriesMomentum}, as opposed to cross-sectional \cite{CrossSectionalMomentum} strategies, which trade assets against each other and select a portfolio based on relative ranking. Volatility scaling \cite{VolTargeting, TSMomAndVolScaling} has been proven to play a crucial role in the positive performance of TSMOM strategies, including deep learning strategies \cite{DeepMomentum}. We scale the returns of each asset by its volatility, so that each asset has a similar contribution to the overall portfolio returns, ensuring that our strategy targets a consistent amount of risk. The consistency over time and across assets has the added benefit of allowing us us to benchmark strategies. Targeting an annualised volatility $\sigma_{\mathrm{tgt}}$, which we take to be $15\%$ in this paper, the realised return of our strategy from day $t$ to $t+1$ is,
\begin{equation}
\label{eqn:tsmom}
R_{t+1}^\mathrm{TSMOM} = \frac{1}{N} \sum_{i=1}^{N} R_{t+1}^{(i)}, \quad R_{t+1}^{(i)} = X_t^{(i)}~\frac{\sigma_{\mathrm{tgt}}}{\sigma_t^{(i)}}~r_{t+1}^{(i)},
\end{equation}
where $X_t$ is our position size, $N$ the number of assets in our portfolio and $\sigma_t^{(i)}$ the ex-ante volatility estimate of the $i$-th asset. We compute $\sigma_t^{(i)}$ using a  60-day exponentially weighted moving standard deviation.

The simplest trading strategy, for which we benchmark performance is \textit{Long Only}, where we always select the maximum position $X^{(i)}_t=1$. The original paper on time-series momentum \cite{TimeSeriesMomentum}, which we will refer to as \textit{Moskowitz}, selects position as $X^{(i)}_t=\sgn(r_{t-252,t})$, where we are using the volatility scaling framework and $r_{t-252,t}$ is annual return. In attempt to react quicker to momentum turning points, \cite{MomentumTurningPoints} blends a slow signal based on annual returns and a fast signal based on monthly returns, to give and \textit{Intermediate} strategy, 
\begin{equation}
    X_t = (1-w)\,\text{sgn}(r_{t-252,t}) + w\,\text{sgn}(r_{t-21,t}).
\end{equation}
We control the relative contribution of the fast and slow signal via $w\in[0,1]$, with the case $w=0$ corresponding to the \textit{Moskowitz} strategy. We additionally use \textit{MACD} \cite{AHLMomentum} as a benchmark, and for details of the implementation we invite the reader to see \cite{DeepMomentum}.

\subsection{Deep Learning}
 We adopt a number of key choices which lead to the improved performance of  DMNs. \\

\subsubsection*{LSTM Architecture} Of the deep-learning architectures assessed in \cite{DeepMomentum}, the Long Short-Term Memory (LSTM) \cite{lstm} architecture yields the best results. LSTM is a special kind of Recurent Neural Network (RNN) \cite{DeepLearningBook}, initially proposed to address the vanishing and exploding gradient problem \cite{vanishinggrad}. An RNN takes an input sequence and, through the use of a looping mechanism where information can flow from one step to another, can be used to transform this into an output sequence while taking into account contextual information in a flexible way. An LSTM operates with cells, which store both short-term memory and long-term memories, using gating mechanisms to summarise and filter information. Internal memory states are sequentially updated with new observations at each step. The resulting model has fewer trainable parameters, is able to learn representations of long-term relationships and typically achieves better generalisation results.\\

\subsubsection*{Trading Signal and Position Sizing} Trading signals are learnt directly by DMNs, removing the need to manually specify both the trend estimator and maps this into a position. The output of the LSTM is followed by a time distributed, fully-connected layer with a activation function $\tanh(\cdot)$, which is a squashing function that directly outputs positions $X^{(i)}_t\in(-1,1)$. The advantage of this approach is that we learn trading rules and positions sizing directly from the data itself. Once our hyperparameters $\theta$ have been trained via backpropagation \cite{backprop}, our LSTM architecture $g(\cdot;\theta)$ takes input features $\textbf{u}^{(i)}_{T-\tau+1:T}$ for all timesteps in the LSTM looking back from time $T$ with $\tau$ steps, and directly outputs a sequence of positions,
\begin{equation}
    \mathbf{X}^{(i)}_{T-\tau+1:T} = g(\mathbf{u}^{(i)}_{T-\tau+1:T}; \theta).
\end{equation}
In an online prediction setting, only the final position in the sequence $X^{(i)}_{T}$ is of relevance to our strategy.\\

\subsubsection*{Loss Function} It has been observed \cite{TrendFollowersLoseMoreThanTheyGain}, that correctly predicting the direction of a stock moves, does not translate directly into a positive strategy return, since the driving moves can often be large but infrequent. Furthermore, we want to account for trade-offs between risk and reward, hence we explicitly optimise networks for risk-adjusted performance metrics. One such metric, used by DMNs is the Sharpe ratio \cite{SharpeRatio}, which calculates the return per unit of volatility. Our Sharpe loss function is,
\begin{equation}
    \label{eqn:sharpe-loss}
    \mathcal{L}_{\mathrm{sharpe}} (\pmb{\theta})  =  -  \frac
    { \sqrt{252} \, \mathbb{E}_\Omega\left[R_t^{(i)}\right] }
    {\sqrt{ \mathrm{Var}_\Omega\left[R_t^{(i)}\right] }},
\end{equation}
where $\Omega$ is the set of all asset-time pairs $\{(i, t)\vert i\in \{1,2,\ldots N\}, t \in \{ T-\tau+1,\ldots,T \} \}$. Automatic differentiation is used to compute gradients for backpropagation \cite{DeepLearningBook}, which explicitly optimises networks for our chosen performance metric.\\

\subsubsection*{Model Inputs} For each timestep, our model can benefit from inputting signals from various timescales. We normalise returns to be $r^{(i)}_{t-t',t}/\sigma_t^{(i)}\sqrt{t'}$, given a time offset of $t'$ days. We use offsets $t'\in\{1,21,63,126,256\}$, corresponding to daily, monthly, quarterly, biannual and annual returns. We also encode additional information by inputting MACD indicators \cite{AHLMomentum}. MACD is a volatility normalised moving average convergence divergence signal, defining the relationship between a short and long signal. For implementation details, please refer to \cite{DeepMomentum}. We use pairs in $\{(8,24), (16,28), (32,96)\}$. We can think of these indicators preforming a similar function to a convolutional layer.

\section{Trading Strategy}
\subsection{Strategy Definition}
As we are using a data-driven approach, we split our training data as a first step, setting aside the first 90\% for training and the last 10\% for validation, for each asset. We calibrate our model using the training data by optimising on the Sharpe loss function \eqref{eqn:sharpe-loss} via minibatch Stochastic Gradient Descent (SGD), using the \textit{Adam} \cite{ADAM} optimiser.  We observe validation loss after each epoch, which is a full pass of the data, to determine convergence. We also use the validation set for the outer optimisation loop, where we tune our model hyperparameters. The hyperparameter optimisation process is detailed in Appendix \ref{apdx:expt-details}. It is necessary to precompute the CPD location $\gamma^{(i)}_t$ and severity $\nu^{(i)}_t$ parameters as detailed by \eqref{eqn:cpd-sev-and-loc}. We do this for each times-asset pair in our training and validation set. It is necessary to do this for a chosen $l\in\{10, 21, 63, 126, 252\}$, corresponding to two weeks, a month, a quarter, half a year and a full year.  We selected these LBW sizes to correspond to input returns timescales, with the exception of the 10 day LBW, which was selected to be as close to daily returns data as reasonably possible. We reinitialise our Matérn 3/2 kernel for each timestep, with all hyperparameters set to $1$. This approach was found to be more stable than borrowing parameters from the previous timestep. For our Changepoint kernel, we initialise the hyperparameters as $c=t-\frac{l}{2}$ and $s=1$. All other parameters are initialised as the equivalent parameter from fitting the  Matérn 3/2 kernel, initialising $k_{\xi_1}$ and $k_{\xi_2}$ with the same values. In the rare case this process fails, we try again by reinitialising all Changepoint kernel parameters to $1$, with the exception of setting $c=t-\frac{l}{2}$. In the event the module still fails for a given timestep, we fill the outputs $\nu^{(i)}_t$ and $\gamma^{(i)}_t$ using the outputs from the previous timestep, noting that we need to increment the changepoint location by an additional step.
 
For each LSTM input, we pass in the normalised returns from the different timescales, our MACD indicators, alongside CPD severity and location, for a chosen $l$. We can either fix $l$ for our strategy or introduce it as a structural hyperparameter, which is tuned by the outer optimisation loop. By doing this, we have information exchange from our CPD Module all the way through to our Sharpe ratio loss function and traded positions. Once our model has been fully trained, we can run it online by computing the CPD module for the most recent data points, then using our LSTM module to select positions to hold for the next day, for each asset.

\subsection{Experiments via Back-testing}
For all of our experiments, we used a portfolio of 50, liquid, continuous futures contracts over the period 1990--2020. The combination of commodities, equities, fixed income and FX futures were selected to make up a well balanced portfolio. The data was extracted from the Pinnacle Data Corp CLC database \cite{PinnacleData}, and the selected futures contracts are listed in Appendix \ref{apdx:data}. All of the selected assets have less than 10\% of data missing. 

In order to back-test our model, we use an expanding window approach, where we start by using 1990--1995 for training/validation, then test out-of-sample on the period 1995--2000. With each successive iteration, we expand the training/validation window by an additional five years, preform the hyperparameter optimisation again, and test on the subsequent five year period. Data was not available from 1990 for every asset and we only use an asset if there is enough data available in the validation set for one at least one LSTM sequence. All of our results are recorded as an average of the test windows. We test our LSTM with CPD strategy using a LBW $l\in\{10, 21, 63, 126, 252\}$, then with the optimised $l$ for each window, based on validation loss.

\begin{figure}[bpth]
\centering
\includegraphics[width=0.9\linewidth]{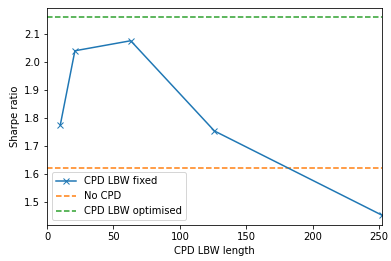}
\label{fig:cpd-window-comparison}
\caption{By increasing the length of the CPD LBW, we observe that even with a short LBW of two weeks, we already start to see performance gains, demonstrating that our GP CPD works well even with limited data. As we continue to increase our CPD window length, performance continues to improve, until it starts to dip after LBW of one quarter. As we approach a LBW of one year, we lose the benefit of the CPD module because it places too much emphasis on larger changepoints which are further in the past. If we introduce LBW as a hyperparameter to be re-evaluated as the training window continues to expand, we observe an additional 4\% increase in Sharpe, leading to a total increase of 33\% over the LSTM baseline.}
\label{fig:varying_cpd}
\end{figure}

We benchmark our strategy against those discussed in Section \ref{sec:mom-strats}, where we choose $w\in\{0,0.5,1\}$ for the \textit{Intermediate} strategy. We also compare our strategy to a DMN which does not have the CPD module. To maintain consistency with the previous work from \cite{DeepMomentum} we benchmark strategy,
\begin{enumerate}
    \item \textbf{profitability} through annualised returns and percentage of positive captured returns,
    \item \textbf{risk} through annualised volatility, annualised downside deviation and maximum drawdown (MDD), and
    \item \textbf{risk adjusted performance} through annualised Sharpe, Sortino and Calmar Ratios.
\end{enumerate}
We provide results for both the raw signal output and then with an additional layer of volatility rescaling to the target of 15\%, for ease of comparison between strategies. It should be noted that this paper selects a more realistic 50 asset portfolio instead of the full 88 assets previously selected in \cite{DeepMomentum}. We focus on raw predictive power of the model and do not account for transaction costs at this stage; however this is a simple adjustment and can easily be incorporated into the loss function. We have included some details and analysis of transaction costs in Appendix \ref{apdx:transaction-costs}. For further information on the implementation and the effects of transaction costs, please refer to \cite{DeepMomentum}.

\begin{table*}[htbp]
\centering
\caption{Strategy Performance Benchmark -- Raw Signal Output.}
\label{tbl:results-raw}
\begin{tabular}{llllllllll}
\hline
\textbf{}                 & \textbf{Returns} & \textbf{Vol.} & \textbf{Sharpe}   & \textbf{\begin{tabular}[c]{@{}l@{}}Downside\\     Deviation\end{tabular}}  & \textbf{Sortino} & \textbf{MDD}    & \textbf{Calmar} & \textbf{\begin{tabular}[c]{@{}l@{}}\% of $+$ve \\     Returns\end{tabular}} & \textbf{$\mathbf{\frac{\text{Ave.  P}}{\text{Ave. L}}}$} \\ \midrule
{\ul \textbf{Reference}} \\
Long Only&2.30\%&5.22\%&0.44&3.59\%&0.64&3.12\%&0.79&52.45\%&0.975\\
MACD&2.65\%&3.58\%&0.77&2.57\%&1.09&2.56\%&0.95&53.34\%&1.002\\ \midrule
{\ul \textbf{TSMOM}} \\
$w=0$&4.41\%&4.80\%&0.94&3.44\%&1.32&3.22\%&1.35&54.28\%&0.990\\
$w=0.5$&3.29\%&3.78\%&0.89&2.80\%&1.23&2.70\%&1.16&53.88\%&0.998\\
$w=1$&2.17\%&4.71\%&0.48&3.29\%&0.68&3.24\%&0.67&51.48\%&1.026\\ \midrule
{\ul \textbf{LSTM}}&3.53\%&2.52\%&1.62&1.71\%&2.46&1.72\%&2.79&55.23\%&1.075\\ \midrule
{\ul \textbf{LSTM w/ CPD}}\\
10-day LBW&3.04\%&1.57\%&1.77&\textbf{1.07\%}&2.74&1.09\%&2.78&55.50\%&1.096\\
21-day LBW&\textbf{3.68}\%&1.81\%&2.04&1.21\%&3.07&1.08\%&\textbf{3.75}&\textbf{56.43}\%&1.095\\
63-day LBW&3.51\%&\textbf{1.72\%}&2.08&1.10\%&3.27&\textbf{1.06\%}&3.58&55.61\%&1.140\\
126-day LBW&3.37\%&2.28\%&1.75&1.59\%&2.66&1.52\%&2.88&54.95\%&1.117\\
252-day LBW&2.81\%&2.24\%&1.45&1.57\%&2.19&1.54\%&2.32&54.00\%&1.101\\
LBW Optimised&3.64\%&1.73\%&\textbf{2.16}&1.17\%&\textbf{3.33}&1.14\%&3.50&56.22\%&\textbf{1.133}\\ \bottomrule
\end{tabular}

\bigskip
\centering
\caption{Strategy Performance Benchmark -- Rescaled to Target Volatility of 15\%.}
\label{tbl:results-rescaled}
\begin{tabular}{llllllllll}
\hline
\textbf{}                 & \textbf{Returns} & \textbf{Vol.} & \textbf{Sharpe}   & \textbf{\begin{tabular}[c]{@{}l@{}}Downside\\     Deviation\end{tabular}}  & \textbf{Sortino} & \textbf{MDD}    & \textbf{Calmar} & \textbf{\begin{tabular}[c]{@{}l@{}}\% of $+$ve \\     Returns\end{tabular}} & \textbf{$\mathbf{\frac{\text{Ave.  P}}{\text{Ave. L}}}$} \\ \midrule
{\ul \textbf{Reference}} \\
Long Only&6.62\%&15.00\%&0.44&10.32\%&0.64&8.96\%&0.79&52.45\%&0.975\\
MACD&11.08\%&15.00\%&0.77&10.74\%&1.09&10.72\%&0.95&53.34\%&1.002\\ \midrule
{\ul \textbf{TSMOM}} \\
$w=0$&13.79\%&15.00\%&0.94&10.74\%&1.32&10.05\%&1.35&54.28\%&0.990\\
$w=0.5$&13.06\%&15.00\%&0.89&11.10\%&1.23&10.72\%&1.16&53.88\%&0.998\\
$w=1$&6.89\%&15.00\%&0.48&10.46\%&0.68&10.32\%&0.67&51.48\%&1.026\\ \midrule
{\ul \textbf{LSTM}}&21.03\%&15.00\%&1.62&10.15\%&2.46&10.24\%&2.79&55.23\%&1.075\\ \midrule
{\ul \textbf{LSTM w/ CPD}}\\
10-day LBW&29.01\%&15.00\%&1.77&10.18\%&2.74&10.39\%&2.78&55.50\%&1.096\\
21-day LBW&30.57\%&15.00\%&2.04&10.06\%&3.07&\textbf{9.01\%}&\textbf{3.75}&\textbf{56.43}\%&1.095\\
63-day LBW&30.71\%&15.00\%&2.08&\textbf{9.65\%}&3.27&9.22\%&3.58&55.61\%&1.140\\
126-day LBW&22.16\%&15.00\%&1.75&10.44\%&2.66&9.99\%&2.88&54.95\%&1.117\\
252-day LBW&18.82\%&15.00\%&1.45&10.54\%&2.19&10.32\%&2.32&54.00\%&1.101\\
LBW Optimised&\textbf{31.52\%}&15.00\%&\textbf{2.16}&10.10\%&\textbf{3.33}&9.88\%&3.50&56.22\%&\textbf{1.133}\\ \bottomrule
\end{tabular}
\end{table*}

\subsection{Results and Discussion}
Our aggregated out-of-sample prediction results, averaged across all five-year windows from 1995--2020, are recorded in Exhibit \ref{tbl:results-raw} and again in Exhibit \ref{tbl:results-rescaled} using volatility rescaling. We plot the effect of CPD LBW size on average Sharpe ratio in Exhibit \ref{fig:varying_cpd} and demonstrate how optimising on this as a hyperparameter can improve overall performance. We note that the CPD computation becomes more intensive for $l\in\{126,252\}$, however we find that performance gains diminish by this point and $l=252$ especially provides no benefit. Impressively, due to our GP framework for CPD, we are able to achieve superior results with very small LBWs, with a notable performance boost from only a two week LBW and performance almost maxes out after only one month. 

Another idea involved passing in outputs from multiple CPD modules with different LBWs in parallel, as inputs to the LSTM. This was not found to improve the model and actually resulted in a degraded performance. It is proposed that multiple LBWs could be useful if using a more complex deep learning architecture than LSTM.

\begin{figure*}[bpth]
\centering
\includegraphics[width=0.95\linewidth]{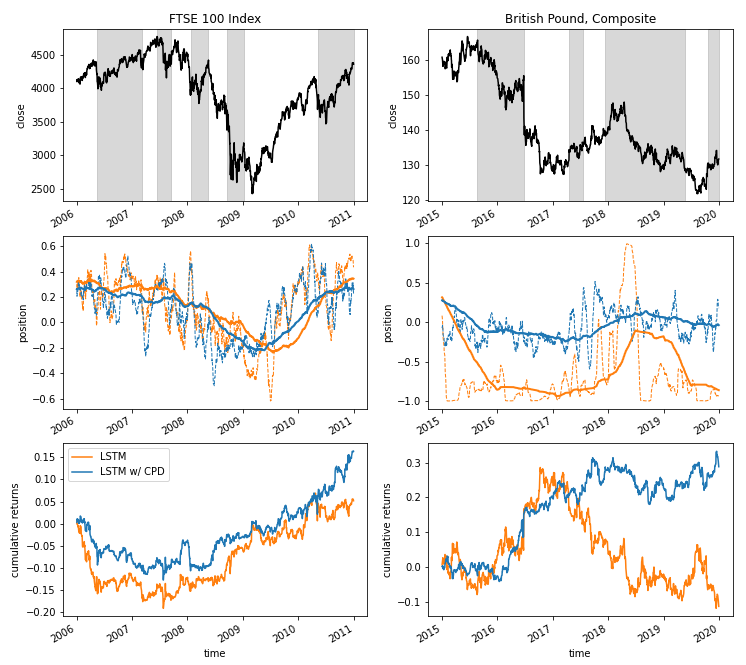}
\caption{These plots examine the positions our DMN takes for single assets, during periods of regime change, providing a comparison of a DMN with and without the CPD module. The top plots track the daily closing price, with the alternating white and grey regions indicating regimes separated by significant changepoints. CPD is performed online with a 63 day LBW, with the changepoint severity $\nu^{(i)}_t \geq 0.9995$ on the left plot and $\nu^{(i)}_t \geq 0.995$ on the right plot indicating a changepoint. Each case uses a 63 day burn in time before we can classify a subsequent changepoint. The middle plots compare the moving averages of position size taken for over a long timescale of one year, indicated by the solid lines, and a shorter timescale of one month, indicated by the dashed line. The bottom plots indicate cumulative returns for each strategy.  The plots on the left look at the FTSE 100 Index during the lead up to the 2008 final crash and its aftermath. With the addition of CPD, our strategy is able to exploit persisting trends with better timing. It is quicker to react to the first dip in 2008 taking short positions to exploit the Bear market with a slow momentum strategy and is similarly able to react to adapt to the Bull market established in 2009 moving to a long strategy quicker. Both approaches exhibit a fast reverting strategy, however after the addition of CPD the strategy is slightly less aggressive with positions taken in response to localised changes. The plots on the right look at the British Pound exchange rate in the lead up to the Brexit vote in 2016 and its aftermath. Here, the Bull and Bear regimes are both less defined and there is a higher level of nonstationarity. With the addition of the CPD module, our model takes a much more conservative slow momentum strategy and instead opts to focus more on achieving positive returns via a fast mean reverting strategy. \label{fig:fast-and-slow}}
\end{figure*}

\begin{figure*}[bpth]
\centering
\begin{subfigure}[]{\linewidth}
\includegraphics[width=1\linewidth]{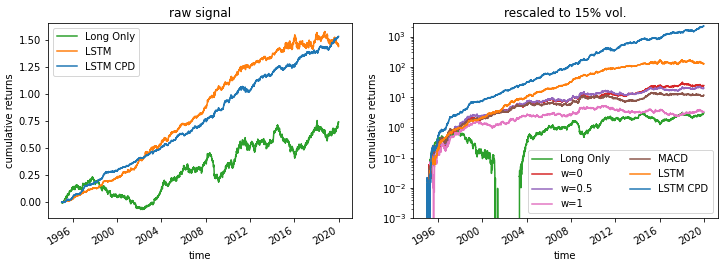}
\end{subfigure}
\caption{These plots benchmark our strategy performance. The plot on the left, of raw signal output, demonstrates that via the introduction of the CPD module we are able to reduce the strategy volatility, especially during the market nonstationarity of more recent years.  With the exception of \textit{Long Only}, we omit the reference strategies in this plot, to avoid clutter. The plot on the right, of signal with rescaled volatility, demonstrates that our strategy outperforms all benchmarks with risk adjusted performance. We show \textit{Intermediate} strategy output for $w\in\{0,0.5,1\}$. \label{fig:strat-benchmark}}
\end{figure*}

In Exhibit \ref{fig:fast-and-slow} we observe how we have slow momentum and fast reversion strategies happening simultaneously. By introducing CPD, we are able to achieve superior returns because we are better able to learn the timing of these strategies and when to place more emphasis on one of these, using a data-driven approach. In our results we can see the difficulties of trying to address regime change with handcrafted techniques such as the \textit{Intermediate} $w=0.5$, which in our experiments actually fails to outperform the $w=0$ \textit{Moskowitz} strategy on all risk adjusted performance ratios.

Our results demonstrate that via the introduction of the CPD module, we outperform the standard DMN in every single performance metric. Our model correctly classifies the direction of the return more often and in addition has a higher average profit to loss ratio. We can see that the CPD module helps to reduce the risk, reducing volatility, downside deviation and MDD, whilst still achieving slightly higher raw returns. This translates to an improvement in risk adjusted performance, improving Sortino ratio by 35\% and Calmar ratio by 25\%. These metrics suggest that the CPD module makes our model more robust to market crashes. We observe an improvement of Sharpe ratio, our target metric, of 33\% which translates to an improvement of 130\% when comparing to the best performing TSMOM strategy. We plot the raw and rescaled signals to benchmark strategies in Exhibit \ref{fig:strat-benchmark}. We note that up until about 2003, when the uptake of electronic trading was becoming much more widespread, the traditional TSMOM and MACD strategies are comparable to the results achieved via the LSTM DMN architecture. At this point the LSTM starts to significantly outperform these traditional strategies, until more recent years where we see volatility increase and performance, especially risk-adjusted performance, drop significantly. This drop in performance can be largely attributed to increased market nonstationarity. Impressively, with the addition of the CPD module, our DMN pipeline continues to perform well even during the market nonstationarity of the 2015--2020 period. Using five repeated trials of the entire experiment, with and without CPD, the average improvement for Sharpe ratio in this period is 70\%, for LBW $l=21$.

\section{Conclusions}
We have demonstrated that the introduction of an online changepoint detection (CPD) module is a simple, yet effective, way to significantly improve model performance, specifically \textit{Deep Momentum Networks} (DMNs). Our model is able blend different strategies at different timescales, learning to do so in a data-driven manner, directly based on our desired risk-adjusted performance metric. In periods of stability, our model is able to achieve superior returns by focusing on slow momentum whilst exploiting but not overreacting to local mean-reversion. The impressive performance increase in periods of nonstationarity, such as recent years, can be attributed to the fact that we 1) can effectively incorporate CPD online with a very short lookback window due to the fact we do so using Gaussian Processes, and 2) pass changepoint score $\nu_t^{(i)}$ from our CPD module to the DMN, helping our model learn how to respond to varying degrees of disequilibrium. As a result, we enhance performance in such conditions where we observe a more conservative slow momentum strategy with a focus on fast mean-reversion.

Future work includes incorporating a CPD module into other deep learning architectures or performing CPD on a model representation as opposed to model inputs. The work in this paper has natural parallels to the field of Continual Learning (CL), which is a paradigm whereby an agent sequentially learns new tasks. Another direction of work will involve utilising CL for momentum trading, where CPD is used to determine task boundaries.

\section{Acknowledgements}
We would like to thank the Oxford-Man Institute of Quantitative Finance for financial and computing support.

\newpage
\bibliographystyle{IEEEtran}
{\footnotesize
\bibliography{mom_bib}
} 
\clearpage
\newpage
\appendix
\subsection{Dataset Details}
\label{apdx:data}

\begin{table}[!htb]
\centering
\begin{tabular}{lll}
\midrule
\textbf{Identifier} & \textbf{Description} & \textbf{\begin{tabular}[c]{@{}l@{}}Back-test\\     From\end{tabular}}       \\ \midrule
{\ul \textbf{Commodities}}\\
CC                  & COCOA                  & 1995    \\ 
DA                  & MILK III, composite      & 2000      \\
GI                  & GOLDMAN SAKS C. I.     & 1995    \\
JO                  & ORANGE JUICE       & 1995        \\
KC                  & COFFEE       & 1995              \\
KW                  & WHEAT, KC        & 1995          \\
LB                  & LUMBER       & 1995              \\
NR                  & ROUGH RICE      & 1995           \\
SB                  & SUGAR \#11    & 1995             \\
ZA                  & PALLADIUM, electronic  & 1995     \\
ZC                  & CORN, electronic   & 1995        \\
ZF                  & FEEDER CATTLE, electronic & 1995 \\
ZG                  & GOLD, electronic       & 1995    \\
ZH                  & HEATING OIL, electronic & 1995   \\
ZI                  & SILVER, electronic      & 1995   \\
ZK                  & COPPER, electronic      & 1995   \\
ZL                  & SOYBEAN OIL, electronic  & 1995  \\
ZN                  & NATURAL GAS, electronic  & 1995  \\
ZO                  & OATS, electronic         & 1995  \\
ZP                  & PLATINUM, electronic    & 1995   \\
ZR                  & ROUGH RICE, electronic   & 1995  \\
ZT                  & LIVE CATTLE, electronic  & 1995  \\
ZU                  & CRUDE OIL, electronic    & 1995  \\
ZW                  & WHEAT, electronic       & 1995   \\
ZZ                  & LEAN HOGS, electronic   & 1995  \\ \hline
{\ul \textbf{Equities}}\\
CA                  & CAC40 INDEX       & 2000         \\
EN                  & NASDAQ, MINI        & 2005       \\
ER                  & RUSSELL 2000, MINI    & 2005     \\
ES                  & S \& P 500, MINI      & 2000     \\
LX                  & FTSE 100 INDEX        & 1995     \\
MD                  & S\&P 400 (Mini electronic) & 1995 \\
SC                  & S \& P 500, composite & 2000     \\
SP                  & S \& P 500, day session  & 1995  \\
XU                  & DOW JONES EUROSTOXX50   & 2005   \\
XX                  & DOW JONES STOXX 50   & 2005      \\
YM                  & Mini Dow Jones (\$5.00) & 2005   \\ \hline
{\ul \textbf{Fixed Income}}\\
DT                  & EURO BOND (BUND)  & 1995      \\
FB                  & T-NOTE, 5yr composite & 1995   \\
TY                  & T-NOTE, 10yr composite & 1995   \\
UB                  & EURO BOBL      & 2005          \\
US                  & T-BONDS, composite  & 1995     \\ \hline
{\ul \textbf{FX}}\\
AN & AUSTRALIAN \$\$, composite & 1995   \\
BN & BRITISH POUND, composite & 1995  \\
CN & CANADIAN \$\$, composite & 1995     \\
DX & US DOLLAR INDEX     & 1995       \\
FN & EURO, composite    & 1995        \\
JN & JAPANESE YEN, composite  & 1995  \\
MP & MEXICAN PESO     & 2000          \\
NK & NIKKEI INDEX  & 1995              \\
SN & SWISS FRANC, composite & 1995     \\ \bottomrule
\end{tabular}
\end{table}

\subsection{Experiment Details}
\label{apdx:expt-details}
We split our data into training and validation datasets using a 90\%, 10\% split. We winsorise our data by limiting it to be within 5 times its exponentially weighted moving (EWM) standard deviations from its EWM average, using a 252-day half life. We calibrate our model using the training data by optimising on the Sharpe loss function via minibatch Stochastic Gradient Descent (SGD), using the \textit{Adam} \cite{ADAM} optimiser. We limit our training to 300 epochs, with an early stopping patience of 25 epochs, meaning training is terminated if there is no decrease in validation loss during this time period. The model is implemented via the \texttt{Keras} API in \texttt{TensorFlow}. Our LSTM sequence length was set to 63 for all experiments. For training and validation, in attempt to prevent overfitting, we split our data into non-overlapping sequences, rather than using a sliding window approach. A stateless LSTM is used, meaning the last state from the previous batch is not used as the initial state for the subsequent batch. Keeping the order of each individual sequence in tact, we shuffle the order which each sequence appears in an epoch. We employ dropout regularisation \cite{dropout} as another technique to avoid overfitting, applying it to LSTM inputs and outputs.

We tune our hyperparameters, with options listed in Exhibit \ref{tab:hyperparams}, using an outer optimisation loop. We achieve this via 50 iterations of random grid search to identify the optimal model. We perform the full experiment for each choice of CPD LBW length and then use the model which achieved the lowest validation loss for the optimised CPD model.

\begin{table}[]
\centering
\caption{Hyperparameter Search Range}
\label{tab:hyperparams}
\begin{tabular}{@{}lll@{}}
\toprule
\textbf{Hyperparameters}           & \textbf{Random Search Grid}                                \\ \midrule
Dropout Rate                       & 0.1, 0.2, 0.3, 0.4, 0.5                                     
\\
Hidden Layer Size                  & 5, 10, 20, 40, 80, 160                                           
\\
Minibatch Size                     & 64, 128, 256 \\
Learning Rate                      & $10^{-4},~ 10^{-3},~ 10^{-2},~ 10^{-1}$                       \\
Max Gradient Norm                  & $10^{-2},~ 10^{0},~ 10^{2}$                        \\
$^*$CPD LBW Length               & 10, 21, 63, 126, 252 \\

\bottomrule
\end{tabular}
\vspace{1ex}

 {\raggedright $\,\,^*$CPD LBW length can be either a hyperparameter or fixed. \par}

\end{table}

\subsection{Transaction costs}
\label{apdx:transaction-costs}
\begin{figure*}[!bpth]
\centering
\begin{subfigure}[]{\linewidth}
\includegraphics[width=1\linewidth]{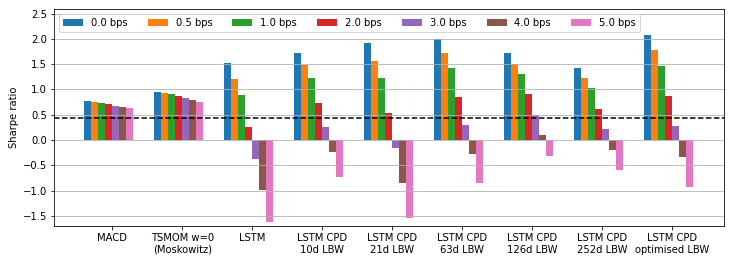}
\end{subfigure}
\caption{A plot of strategy average Sharpe ratio across all five year windows from 1995--2020. We look at the effect of increasing average transaction cost $C$ from $0$ to $5$ basis points (bps), on the Sharpe ratio of our raw signal output. It should be noted that, here, we do not not incorporate $C$ into the loss function. The black dotted line indicates the \textit{Long Only} reference. \label{fig:transaction-costs}}
\end{figure*}
Assuming an average transaction cost of $C$, we calculate turnover adjusted returns as,
\begin{equation}
    \bar{R}_{t+1}^{(i)} 
    = R_{t+1}^{(i)}
    + - C\sigma_{\mathrm{tgt}}
    \left| \frac{ X_t^{(i)}}{\sigma_t^{(i)}} - \frac{ X_{t-1}^{(i)}}{\sigma_{t-1}^{(i)}} \right|
\end{equation}
In Exhibit \ref{fig:transaction-costs} we demonstrate the effects of transaction cost on our raw signal. Our strategy outperforms classical strategies for transaction costs of up to 2 basis points, at which point it rapidly deteriorates, due to the fast reverting component.  We note that the a larger CPD LBW window size becomes favourable as we increase $C$. We suspect this is because the model focuses on larger long term changepoints and favours slow momentum over fast reversion. For larger average transaction costs greater than 1bps we suggest incorporating turnover adjusted returns into the loss function \eqref{eqn:sharpe-loss}. This adjustment is detailed in \cite{DeepMomentum}, where it is demonstrated to work well when transaction costs are high.

\end{document}